\documentclass[10pt,journal,compsoc]{IEEEtran}
\ifCLASSOPTIONcompsoc
  \usepackage[nocompress]{cite}
\else
  \usepackage{cite}
\fi

\usepackage{times}
\usepackage{epsfig}
\usepackage{graphicx}
\usepackage{amsmath}
\usepackage{amssymb}
\usepackage{multirow}
\usepackage{subfig}
\usepackage{url}
\usepackage{tablefootnote}
\usepackage{comment}
\usepackage{epstopdf}
\graphicspath{{./images/}}
\usepackage{refcount}
\def\tablefootnotemark#1{\textsuperscript{\getrefnumber{#1}}}

\begin{document}

\title{Recognizing Disguised Faces in the Wild}

\author{Maneet~Singh,~\IEEEmembership{Student Member,~IEEE,} Richa~Singh,~\IEEEmembership{Senior Member,~IEEE,} Mayank~Vatsa,~\IEEEmembership{Senior Member,~IEEE,} Nalini~Ratha,~\IEEEmembership{Fellow,~IEEE,}~and~ Rama~Chellappa,~\IEEEmembership{Fellow,~IEEE}
\IEEEcompsocitemizethanks{\IEEEcompsocthanksitem M. Singh, R. Singh, and M. Vatsa are with IIIT-Delhi, India, 110020 (e-mail: maneets@iiitd.ac.in, rsingh@iiitd.ac.in, mayank@iiitd.ac.in).
\IEEEcompsocthanksitem N. Ratha is with IBM TJ Watson Research Center, New York, USA (e-mail: ratha@us.ibm.com).
\IEEEcompsocthanksitem R. Chellappa is with Department of Electrical and Computer Engineering, University of Maryland, College Park, MD,
20742 (e-mail: rama@umiacs.umd.edu).
\IEEEcompsocthanksitem DFW dataset link: http://iab-rubric.org/resources/dfw.html \protect \\
Shorter version of this paper paper was presented at CVPR Workshop on DFW, 2018 \cite{dfw}}
}

\markboth{ }%
{Singh \MakeLowercase{\textit{et al.}}: Recognizing Disguised Faces in the Wild}

\IEEEtitleabstractindextext{
\begin{abstract}
Research in face recognition has seen tremendous growth over the past couple of decades. Beginning from algorithms capable of performing recognition in constrained environments, the current face recognition systems achieve very high accuracies on large-scale unconstrained face datasets. While upcoming algorithms continue to achieve improved performance, a majority of the face recognition systems are susceptible to failure under disguise variations, one of the most challenging covariate of face recognition. In literature, some algorithms demonstrate promising results on existing disguise datasets, however, most of the disguise datasets contain images with limited variations, often captured in controlled settings. This does not simulate a real world scenario, where both intentional and unintentional unconstrained disguises are encountered by a face recognition system. In this paper, a novel Disguised Faces in the Wild (DFW) dataset is proposed which contains over 11,000 images of 1,000 identities with variations across different types of disguise accessories. The dataset is collected from the Internet, resulting in unconstrained face images similar to real world settings. This is the first-of-a-kind dataset with the availability of \textit{impersonator} and genuine \textit{obfuscated} face images for each subject. The proposed DFW dataset has been analyzed in terms of three levels of difficulty: (i) easy, (ii) medium, and (iii) hard in order to showcase the challenging nature of the problem. It is our view that the research community can greatly benefit from the DFW dataset in terms of developing algorithms robust to such adversaries. The proposed dataset was released as part of the First International Workshop and Competition on Disguised Faces in the Wild at the International Conference on Computer Vision and Pattern Recognition, 2018. This paper presents the DFW dataset in detail, including the evaluation protocols, baseline results, performance analysis of the submissions received as part of the competition, and three levels of difficulties of the DFW challenge dataset.  
\end{abstract}

}

\maketitle
\IEEEdisplaynontitleabstractindextext
\IEEEpeerreviewmaketitle

\IEEEraisesectionheading{\section{Introduction}}
\IEEEPARstart{E}{xtensive} research in the domain of face recognition has resulted in the development of algorithms achieving state-of-the-art performance on large-scale unconstrained datasets \cite{lfwBest,goswami2017face}. However, it has often been observed that most of these systems are susceptible to digital and physical adversaries \cite{goswami18AAAI,akhtar18Access,galbally14survey,juefei2015preliminary,disguise,disguise2}. Digital adversaries refer to manipulations performed on the image being provided to the recognition system, with the intent of \textit{fooling} the system. It has been shown that traditional systems based on hand crafted features \cite{goswami18AAAI} do well with digital attacks while deep learning systems deteriorate rapidly with digital attacks. Recently, the issue of digital attacks has garnered attention, with perturbation techniques such as Universal Adversarial Perturbation \cite{universal} and DeepFool \cite{deepFool} demonstrating high adversarial performance on different algorithms. On the other hand, physical adversaries refer to the variations brought to the individual before capturing the input data for the recognition system. In case of face recognition, this can be observed due to variations caused by different spoofing techniques or disguises. While the area of spoof detection and mitigation is being well explored \cite{galbally14survey,ramachandra17}, research in the domain of disguised face recognition is yet to receive dedicated attention, despite its significant impact on both traditional and deep learning systems \cite{dfw,tejas}.

\begin{figure}
\centering
\includegraphics[width=3.6in]{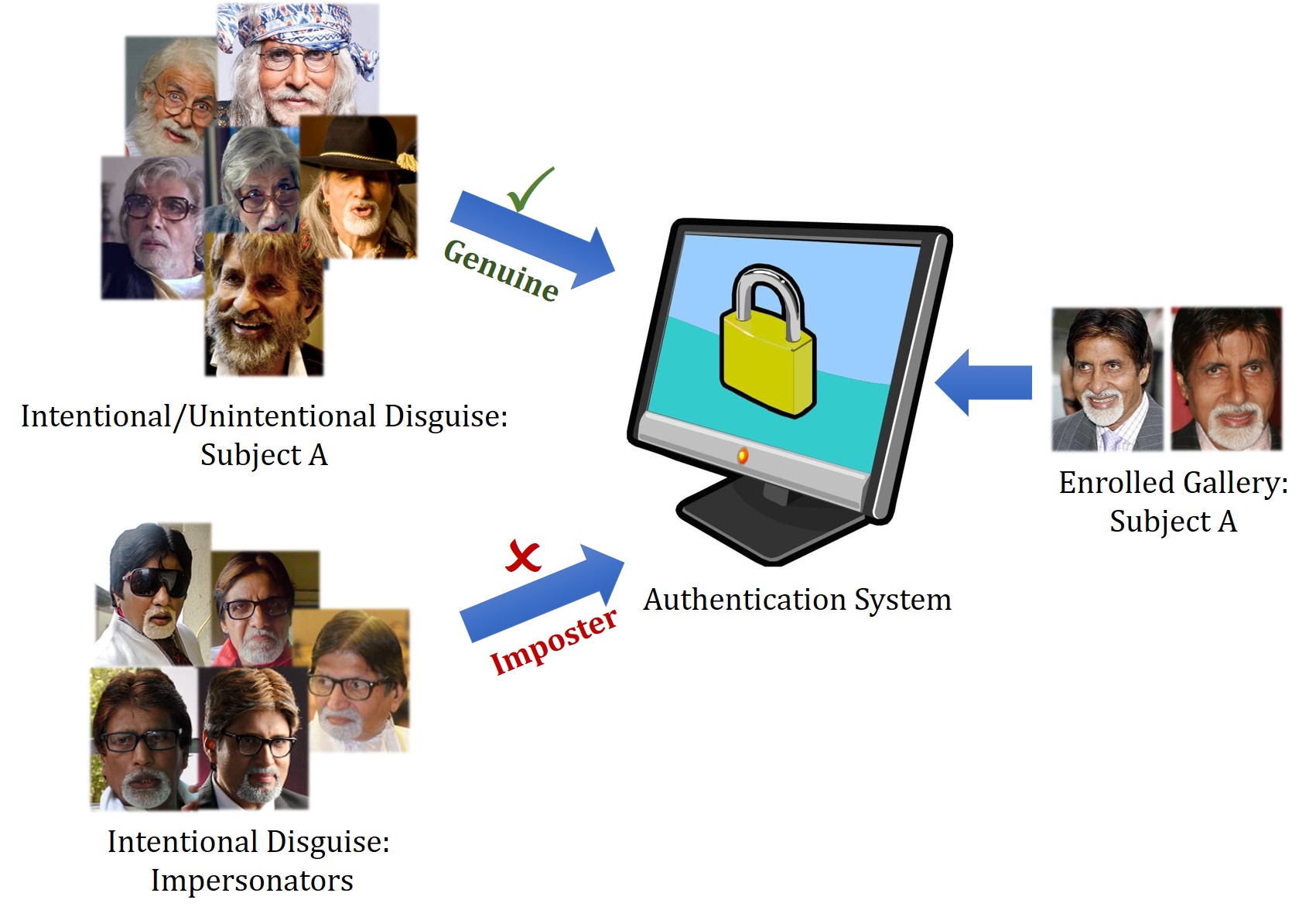}
\caption{Authentication systems often face the challenge of matching disguised face images with non-disguised enrolled images. }
\label{fig:intro}
\vspace{-10pt}
\end{figure}

\renewcommand{\arraystretch}{1.2}
\begin{table*}
\centering
\caption{Summary of disguise face datasets in literature.}
\begin{tabular}{|l|c|c|c|c|c|}
\hline
\multirow{2}{*}{\textbf{Name}} & {\textbf{Controlled}} & \multicolumn{2}{c|}{\textbf{Number of}}  & \textbf{Availability of} & \textbf{Publicly}\\
\cline{3-4}
& \textbf{Disguise} & \textbf{Images} & \textbf{Subjects} & \textbf{Impersonators}  & \textbf{Available} \\
\hline
\hline
AR Dataset (1998) \cite{ar}  & Yes & 3,200 & 126 & No & Yes\\
\hline
National Geographic Dataset (2004) \cite{national}  & Yes & 46 & 1 & No & No\\
\hline
Synthetic Disguise Dataset (2009) \cite{synthetic} & Yes &  4,000 & 100 & No & No \\
\hline
Curtin Faces Dataset (2011) \cite{curtin} & Yes & ~5,000 & 52 & No & Yes \\
\hline
IIITD I$^2$BVSD Dataset (2014) \cite{tejasicb}  & Yes & 1,362 & 75 & No & Yes\\
\hline
Disguised and Makeup Faces Dataset (2016) \cite{hongKong} & No & 2,460 & 410 & No & Yes \\
\hline
Spectral Disguise Face Dataset (2018) \cite{raghavendraDisguise} & Yes & 6,480 & 54 & No & Yes   \\
\hline
\textbf{Proposed DFW Dataset (2018)} & \textbf{No} & \textbf{11,157} & \textbf{1,000} & \textbf{Yes} & \textbf{Yes}\\
\hline
\end{tabular}
\label{tab:dbs}
\vspace{-10pt}
\end{table*}

Disguised face recognition encompasses handling both \textit{intentional} and \textit{unintentional} disguises. Intentional disguise refers to the scenario where a person attempts to hide their identity or \textit{impersonate} another person's identity, in order to fool a recognition system into obtaining unauthorized access. This often results in utilizing external disguise accessories such as wigs, beard, hats, mustache, and heavy makeup, leading to obfuscation of the face region. This renders low inter-class variations between different subjects, thereby making the problem challenging in nature. Unintentional disguises cover a range of images wherein the face is obfuscated by means of an accessory such as glasses, hats, or masks in an unintentional manner. It can also be an effect of aging, resulting in an increase or decrease of facial hair such as beard or mustache, and variations in the skin texture. Unintentional disguise creates challenges for the face recognition system by increasing the intra-class variations for a given subject. The combination of both intentional and unintentional disguises render the problem of disguised face recognition an arduous task. Fig. \ref{fig:intro} presents sample images of intentional and unintentional disguises, along with non-disguised enrolled face images. The authentication system faces the challenge of verifying an image containing unconstrained disguise variations against a frontal non-disguised face image. 

This paper presents a novel Disguised Faces in the Wild (DFW) dataset, containing 11,157 face images of 1,000 identities. Almost the entire dataset is collected from the Internet resulting in an unconstrained set of images. One of the key highlights of the proposed dataset is the availability of (i) \textit{normal}, (ii) \textit{validation}, (iii) \textit{disguised}, and (iv) \textit{impersonator} images for a given subject. This is a first-of-a-kind dataset containing multiple types of in-the-wild images for a subject in order to evaluate different aspects of disguised face recognition, along with three pre-defined evaluation protocols. It is our assertion that the availability of a large-scale dataset, containing images captured in unconstrained settings across multiple devices, pose, illumination, and disguise accessories would help in encouraging research in this direction. The dataset was released as part of the DFW challenge, in the Disguised Faces in the Wild Workshop at International Conference on Computer Vision and Pattern Recognition (CVPR), 2018. We present the DFW dataset, along with the findings across the three evaluation protocols. Performance of participants in the DFW challenge along with the baseline results, and analysis of three difficulty levels has also been provided. The organization of this paper is as follows: Section \ref{sec:background} presents the motivation of the DFW workshop and challenge, followed by a detailed description of the DFW dataset. Section \ref{sec:comp} elaborates upon the DFW challenge, its submissions, and performance across the three protocols. Section \ref{sec:deg} presents the DFW dataset's three degree of difficulties in terms of \textit{easy}, \textit{medium}, and \textit{hard}. 

\section{Motivation}
\label{sec:background}
Table \ref{tab:dbs} presents the characteristics of existing disguise face datasets, along with the proposed DFW dataset. One of the initial datasets containing disguise variations is the AR dataset \cite{ar}. It was released in 1998 and contains 3,200 face images having controlled disguise variations. This was followed by the release of different datasets having variations across the disguise accessories and dataset size. As observed from Table \ref{tab:dbs}, most of the datasets are moderately sized having controlled disguise variations. Other than disguised face datasets, a lot of recent research in face recognition has focused on large-scale datasets captured in unconstrained environments \cite{lfw,megaface,ytf,vggFace2,mf2}. The availability of such datasets facilitate research in real world scenarios, however, they do not focus on the aspect of disguised face recognition.

Disguised face recognition presents the challenge of matching faces under both intentional and unintentional distortions. It is interesting to note that both forms of disguise can result in either genuine or imposter pairs. For instance, a criminal may intentionally attempt to conceal his identity by using external disguise accessories, thereby resulting in a genuine match for an authentication system. On the other hand, an individual might intentionally attempt to impersonate another person, resulting in an imposter pair for the face recognition system. Similarly, in case of unintentional disguises, use of casual accessories such as sunglasses or hats result in a genuine disguised pair, while individuals which look alike are imposter pairs for the recognition system. The combination of different disguise forms along with the intent makes the given problem further challenging.

\begin{table}
\centering
\caption{Statistics of the proposed Disguised Faces in the Wild (DFW) dataset.}
\begin{tabular}{|l|c|}
\hline
\textbf{Characteristic} & \textbf{Count} \\
\hline
\hline
{Subjects} & 1,000 \\
\hline
{Images} & 11,157 \\
\hline
{Normal Images} & 1,000 \\
\hline
{Validation Images} & 903 \\
\hline
{Impersonator Images} & 4,440 \\
\hline
{Range of Images per Subject} & [5,26] \\
\hline
\end{tabular}
\label{tab:dbDetails}
\vspace{-10pt}
\end{table}

To the best of our knowledge, no existing disguise dataset captures the wide spectrum of intentional and unintentional disguises. To this effect, we prepared and released the DFW dataset. The DFW dataset simulates the real world scenario of unconstrained disguise variations, and provides multiple impersonator images for almost all subjects. The presence of impersonator face images enables the research community to analyze the performance of face recognition models under physical adversaries. The dataset was released as part of the DFW workshop, where researchers from all over the World were encouraged to evaluate their algorithms against this challenging task. Inspired by the presence of disguise intent in real world scenarios, algorithms were evaluated on three protocols: (i) Impersonation, (ii) Obfuscation, and (iii) Overall. Impersonation focuses on disguise variations where an individual either attempts to impersonate another individual intentionally, or looks like another individual unintentionally. In both cases, the authentication system should be able to detect an (imposter) unauthorized access attempt. The second protocol, obfuscation, focuses on intentional or unintentional disguise variations across genuine users. In this case, the authentication system should be able to correctly identify genuine users even under varying disguises. The third protocol evaluates a face recognition model on the entire DFW dataset. As mentioned previously, it is our hope that the availability of DFW dataset along with the three pre-defined protocols would enable researchers to develop state-of-the-art algorithms robust to different physical adversaries.

\section{Disguised Faces in the Wild (DFW) Dataset}


As shown in Table \ref{tab:dbs}, most of the research in the field of disguised face recognition has focused on images captured in controlled settings, with limited set of accessories. In real world scenarios, the problem of disguised face recognition extends to data captured in uncontrolled settings, with large variations across disguise accessories. Combined with the factor of \textit{disguise intent}, the problem of disguise face recognition is often viewed as an exigent task. The proposed DFW dataset simulates the above challenges by containing 11,157 face images belonging to 1,000 identities with uncontrolled disguise variations. It is the first dataset which also provides impersonator images for a given subject. The DFW dataset contains the IIIT-Delhi Disguise Version 1 Face Database (ID V1) \cite{tejas} having 75 subjects, and images corresponding to the remaining 925 subjects have been taken from the Internet. Since the images have been taken from the Web, most of the images correspond to famous personalities and encompass a wide range of disguise variations. The dataset contains images with respect to unconstrained disguise accessories such as hair-bands, masks, glasses, sunglasses, caps, hats, veils, turbans, and also variations with respect to hairstyles, mustache, beard, and make-up. Along with the disguise variations, the images also demonstrate variations across illumination, pose, expression, background, age, gender, and camera quality. The dataset is publicly available for research purposes and can be downloaded from our website \footnote{http://iab-rubric.org/resources/dfw.html}. The following subsections present the dataset statistics, protocols for evaluation, and details regarding data distribution. 

\subsection{Dataset Statistics}
As mentioned previously, the proposed DFW dataset contains images pertaining to 1,000 identities, primarily collected from the Internet. Most of the subjects are adult famous personalities of Caucasian or Indian ethnicity. Each subject contains at least 5 and at most 26 images. The dataset comprises of 11,157 face images including different kinds of images for a given subject, that is, \textit{normal}, \textit{validation}, \textit{disguised}, and \textit{impersonator}. Detailed description of each type is given below:

\begin{figure*}
\centering
\subfloat[][Subject A]{\includegraphics[width=6in]{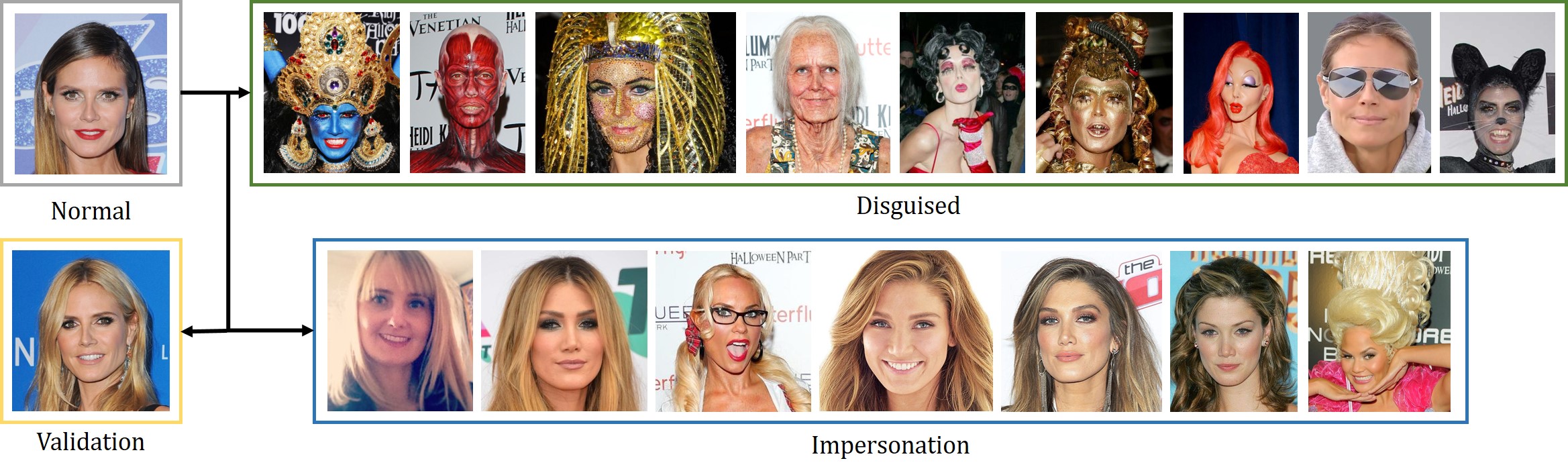}} \\
\subfloat[][Subject B]{\includegraphics[width=6.1in]{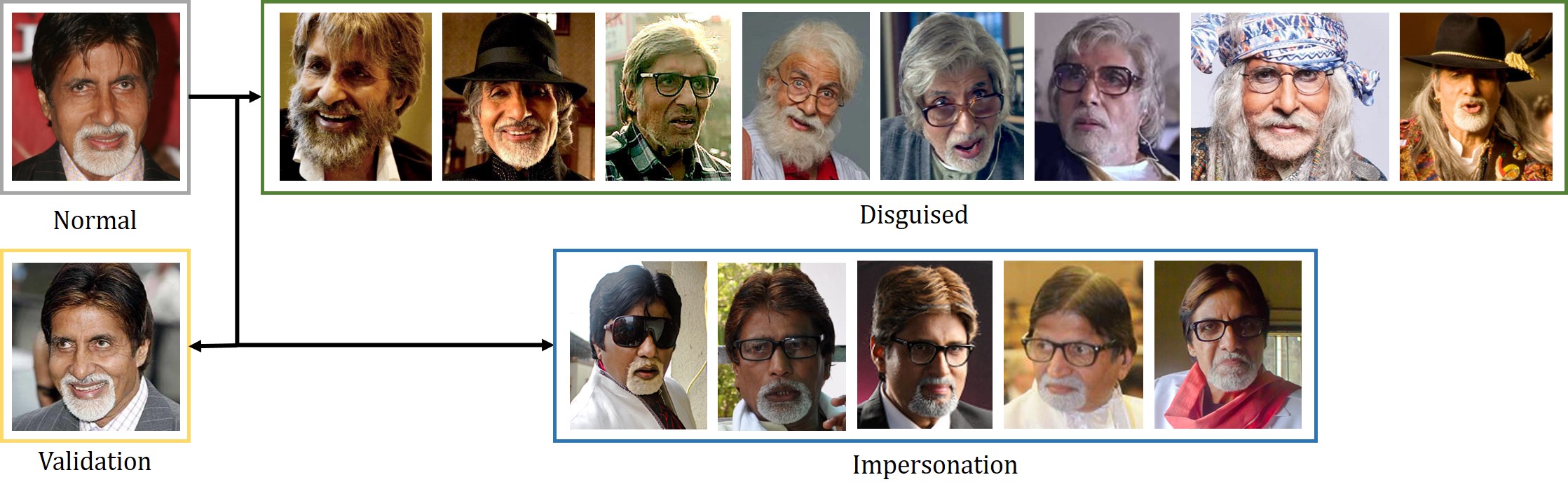}}
\caption{Images pertaining to two subjects of the DFW dataset. The dataset contains at most four types of images for each subject: Normal, Validation, Disguised, and Impersonator.}
\label{fig:db}
\vspace{-10pt}
\end{figure*}

\begin{itemize}
\item \textbf{Normal Face Image:} Each subject has a frontal, non-disguised, good quality face image, termed as the normal face image.
 
\item \textbf{Validation Face Image:} Other than the normal face image, 903 subjects have another non-disguised face image, referred to as the validation image. This can help in evaluating a proposed model for matching non-disguised face images.

\item \textbf{Disguised Face Image:} For each subject, disguised face images refer to images having intentional or unintentional disguise of the same subject. For the 1,000 identities present in the dataset, every subject has at least one and at most 12 disguised images. These images form genuine pairs with the normal and validation face images, and can help in evaluating the true positive rate of an algorithm.

\item \textbf{Impersonator Face Image:} Impersonators refer to people who intentionally or unintentionally look similar to a another person. For a given subject, impersonator face images belong to different people, thereby resulting in imposter pairs which can be used to evaluate the true negative rate of an algorithm. In the DFW dataset, 874 subjects have images corresponding to their impersonators, each having at least 1 and at most 21 images. 
\end{itemize}

Statistics of the proposed dataset are presented in Table \ref{tab:dbDetails}, and Fig. \ref{fig:db} demonstrates sample images of two subjects. It can be observed that disguised face images result in increased intra-class variations for a given subject, while the impersonator images render lower inter-class variability. Overall, the DFW dataset contains 1,000 and 903 normal and validation face images, respectively, 4,814 disguised face images, and 4,440 impersonator images. 


\begin{table}
\centering
\caption{Statistics of the training and testing sets of the DFW dataset.}
\begin{tabular}{|l|c|c|}
\hline
\textbf{Number of} & \textbf{Training Set} & \textbf{Testing Set}\\
\hline
\hline
Subjects & 400 & 600 \\
\hline
Images & 3,386 & 7,771 \\
\hline
Normal Images & 400 & 600 \\
\hline
Validation Images & 308 & 595  \\
\hline
Disguised Images & 1,756 & 3,058  \\
\hline
Impersonator Images & 922 & 3,518 \\
\hline 
\end{tabular}
\label{tab:protocol}
\vspace{-10pt}
\end{table}

\subsection{Protocols for Evaluation}
\label{protocol}
The DFW dataset has been released with three protocols for evaluation. A fixed training and testing split is provided which ensures mutual exclusion of images and subjects. 400 subjects are used to create the training set, and the remaining 600 subjects form the test set. Table \ref{tab:protocol} presents the statistics of the testing and training sets. All three protocols correspond to verification protocols, where a face recognition module is expected to classify a pair of images as genuine or imposter. Detailed description of each protocol on the pre-defined training and testing partitions is given below:

\noindent \textbf{Protocol-1 (Impersonation)} evaluates a face recognition model for its ability to distinguish impersonators from genuine users with high precision. A combination of a normal image with a validation image of the same subject corresponds to a genuine pair for this protocol. For imposter pairs, the impersonator images of a subject are partnered with the normal, validation, and disguised images of the same subject.

\noindent \textbf{Protocol-2 (Obfuscation)} is useful for evaluating the performance of a face recognition system under intentional or unintentional disguises, wherein a person attempts to hide his/her identity. The genuine set contains pairs corresponding to the (normal, disguise), (validation, disguise), and (disguise$_1$, disguise$_2$) images of a subject. Here, disguise$_n$ corresponds to the $n^{th}$ disguised image of a subject. That is, all pairs generated using the normal and validation images with the disguise images, and the pairs generated between the disguise images of the same subject, constitute the genuine pairs. The imposter set is created by combining the normal, validation, and disguised images of one subject with the normal, validation, and disguised images of a different subject. This results in the generation of cross-subject imposter pairs. The impersonator images are not used in this protocol.

\noindent \textbf{Protocol-3 (Overall Performance)} is used to evaluate the performance of any face recognition algorithm on the entire DFW dataset. The genuine and imposter sets created in the above two protocols are combined to generate the data for this protocol. For the genuine set, pairs are created using the (normal, validation), (normal, disguise), (validation, disguise), and (disguise$_1$, disguise$_2$) images of the same subject. For the imposter set, cross-subject imposter pairs are considered, wherein the normal, validation, and disguised face images of one subject are combined with the normal, validation, and disguised face images of another subject. Apart from the cross-subject imposter pairs, the impersonators of one subject are also combined with the normal, validation, and disguised face images of the same subject to further supplement the imposter set.

\begin{table*}
\centering
\caption{List of teams which participated in the DFW competition.}
\begin{tabular}{|l|l|l|}
\hline
\textbf{Model} & \textbf{Affiliation} & \textbf{Brief Description} \\
\hline
\hline
\multirow{2}{*}{AEFRL \cite{aefrlItmo}} &  \ \multirow{2}{*}{\shortstack[l]{The Saint-Petersburg National Research University of Information \\ Technologies, Mechanics and Optics (ITMO), Russia}} & MTCNN + 4 networks for feature extraction + Cosine distance \\ 
& & \\ 
\hline
ByteFace & Bytedance Inc., China & Weighted linear combination of ensemble of 3 CNNs \\
\hline
DDRNET \cite{ddrnetWvu}& West Virginia University, USA & Inception Network with Center Loss \\
\hline
DisguiseNet \cite{disguiseNetRopar} & Indian Institute of Technology Ropar, India & Siamese network with VGG-Face having a weighted loss\\
\hline
DR-GAN & Michigan State University, USA & MTCNN + DR-GAN + Cosine distance\\
\hline
LearnedSiamese & Computer Vision Center UAB, Spain & Cropped faces + Siamese Neural Network \\
\hline
MEDC & Northeastern University, USA & MTCNN + Ensemble of 3 CNNs + Average Cosine distance \\
\hline
MiRA-Face \cite{miraFaceNTU} & National Taiwan University, Taiwan & MTCNN + RSA + Ensemble of CNNs \\
\hline
OcclusionFace & Zhejiang University, China & MTCNN + Fine-tuned ResNet-28 \\
\hline
Tessellation & Tessellate Imaging, India & Siamese network with triplet loss model \\
\hline
UMDNets \cite{resnetUmd} & The University of Maryland, USA & All-In-One + Average across scores obtained by 2 networks \\
\hline
WVU\_CVL & West Virginia University, USA & MTCNN + CNN + Softmax \\
\hline
\end{tabular}
\label{tab:teams}
\end{table*}

\subsection{Nomenclature and Data Distribution}
The DFW dataset is available for download as an archived file containing one folder for each subject. Each of the 1,000 folders is named with the subject's name and may contain the four types of images discussed above: normal, validation, disguise, and impersonator. In order to ensure consistency and eliminate ambiguity, the following nomenclature has been followed across the dataset:

\begin{itemize}
\item Each subject has a single normal face image, which has been named as \textit{firstName\_lastName.jpg}. For instance, for the subject Alicia Keys, the subject's normal image is named \textit{Alicia\_Keys.jpg}.

\item As mentioned previously, a given subject contains only a single validation face image. Therefore, the validation image is named with a postfix `\_a', that is, \textit{firstName\_lastName\_a.jpg}. For the example of Alicia Keys, the subject validation image is stored as \textit{Alicia\_Keys\_a.jpg}.

\item For disguised face images, a postfix of `\_h' is adopted, along with a number for uniquely identifying the disguised face image of a given subject. That is, \textit{firstName\_lastName\_h\_number.jpg}. Here, \textit{number} can take values such as `001', `002', ... `010'. For example, the first disguise image of subject Alicia Keys can be named as \textit{Alicia\_Keys\_h\_001.jpg}, while the third disguised face image can be named as \textit{Alicia\_Keys\_h\_003.jpg}.

\item Similar to the disguised image nomenclature, a postfix of `\_I' is used to store the impersonator images of a subject. That is, impersonator images are named as \textit{firstName\_lastName\_I\_number.jpg}. For example, the first impersonator image of subject Alicia Keys can be named as \textit{Alicia\_Keys\_I\_001.jpg}.
\end{itemize}

In order to correctly follow the protocols mentioned above, and report corresponding accuracies, training and testing mask matrices are also provided along with the dataset. Given the entire training or testing partition, the mask matrix can be used to extract relevant genuine and imposter pairs or scores for a given protocol. The DFW dataset also contains face co-ordinates obtained via faster RCNN \cite{fasterRcnn}. Given an image of the dataset, the co-ordinates provide the face location in the entire image. 

\section{Disguised Faces in the Wild Competition}
\label{sec:comp}
Disguised Faces in the Wild competition was conducted as part of the \textit{First International Workshop on Disguised Faces in the Wild}\footnote{http://iab-rubric.org/DFW/dfw.html}, at the International Conference on Computer Vision and Pattern Recognition, 2018 (CVPR'18). Participants were required to develop a disguised face recognition algorithm, which was evaluated on all three protocols of the DFW dataset. The competition was open world-wide, to both industry and academic institutions. The competition saw over 100 registrations from across the world. 

All participating teams were provided with the DFW dataset, including the training and testing splits, face co-ordinates, and mask matrices for generating the genuine and imposter pairs. Evaluation was performed based on the three protocols described in Section \ref{protocol}. No restriction was enforced in terms of utilizing external training data, except ensuring mutual exclusion with the test set. 
The remainder of this section presents the technique and performance analysis of all the submissions, including the baseline results.  

\subsection{Baseline Results}
Baseline results are computed using the VGG-Face descriptor \cite{vgg}, which is one of the top performing deep learning models for face recognition. A pre-trained VGG-Face model is used for feature extraction (trained on the VGG-Face dataset \cite{vgg}). The extracted features are compared using Cosine distance, followed by classification into genuine or imposter. Baseline results were also provided to the participants. 

\subsection{DFW Competition: Submissions}
The DFW competition received 12 submissions from all over the world, having both industry and academic affiliations. Table \ref{tab:teams} presents a list of the participating teams, along with their affiliation. Details regarding the technique applied by each submission is provided below: \\

\begin{figure*}
\centering
\subfloat[AEFRL]{\includegraphics[height=2.2in]{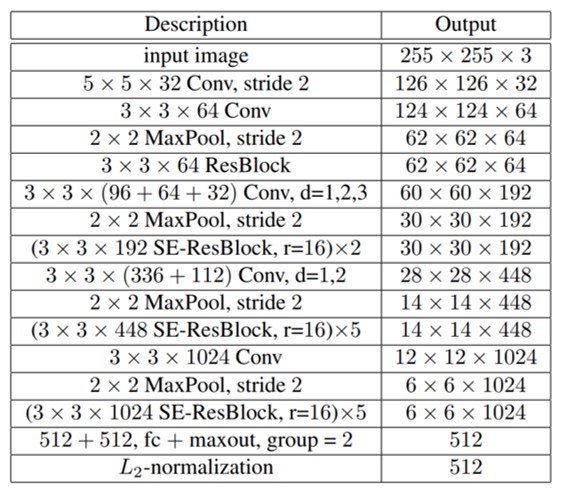}}
\hspace{1em}
\subfloat[UMDNets]{\includegraphics[height=2.2in]{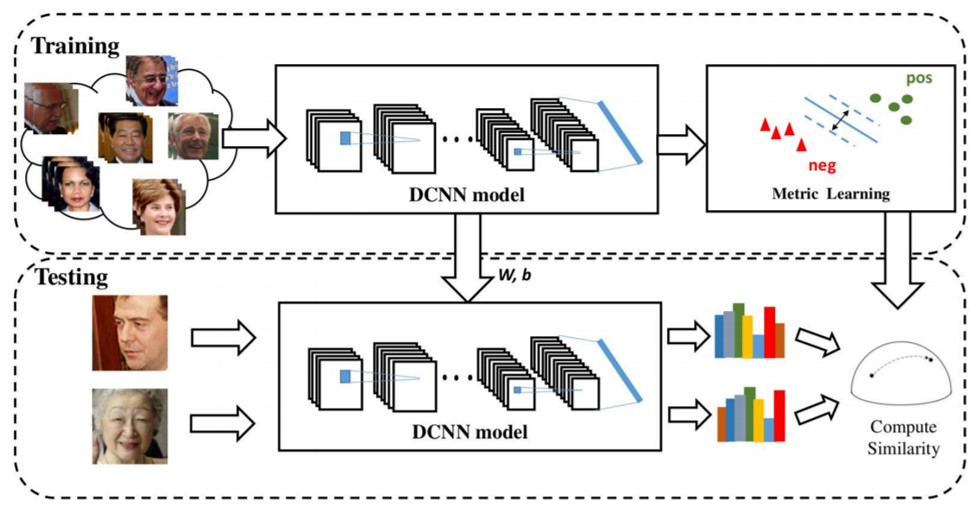}}
\caption{Diagrammatic representation of (a) AEFRL \cite{aefrlItmo}, and (b) UMDNets \cite{resnetUmd}. Images have been taken from their respective publications.}
\label{fig:arch1}
\end{figure*}

\noindent\textbf{(i) Appearance Embeddings for Face Representation Learning (AEFRL) \cite{aefrlItmo}:} AEFRL is a submission from the Information Technologies, Mechanics and Optics (ITMO), Russian Federation. Later in the competition, it was renamed to Hard Example Mining with Auxiliary Embeddings. Faces are detected, aligned, and cropped using Multi-task Cascaded Convolutional Networks (MTCNN) \cite{mtcnn}. This is followed by horizontal flipping, and feature extraction by four separate networks. Feature-level fusion is performed by concatenation of the features obtained for the original and flipped image, followed by concatenation of all features from different networks. $l_2$ normalization is performed on the concatenated feature vector, followed by classification using Cosine distance. The CNN architecture used in the proposed model is given in Fig. \ref{fig:arch1}(a).
\vspace{3pt}

\begin{figure}
\centering
\includegraphics[width=3.5in]{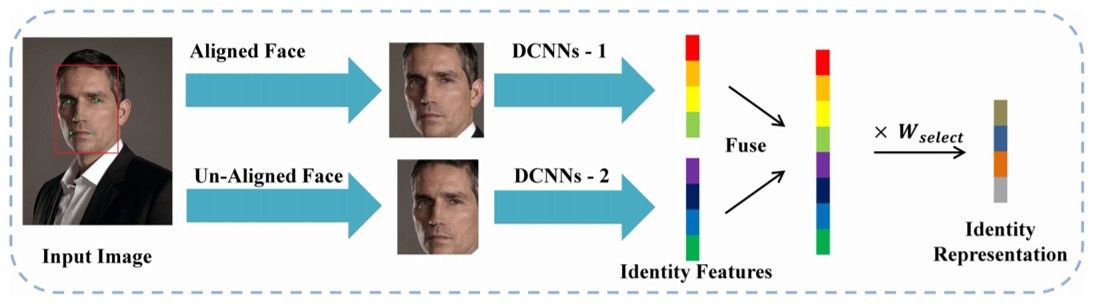}
\caption{Diagrammatic representation of MiRA-Face \cite{miraFaceNTU}. Image has directly been taken from their publication. }
\label{fig:arch2}
\end{figure}

\noindent\textbf{(ii) ByteFace:} Proposed by a team from Bytedance Inc., China, ByteFace uses an ensemble of three CNNs for performing disguised face recognition. For detection and alignment, the algorithm uses a mixture of co-ordinates provided with the DFW dataset and MTCNN. Three CNNs are trained with (i) modified center loss and Cosine similarity \cite{center}, (ii) joint Bayesian similarity, and (iii) sphere face loss \cite{sphere} with joint Bayesian similarity, respectively. A linear weighted combination of scores obtained via the three models is used for performing the final classification. The CASIA WebFace \cite{webface} dataset is also used for training the proposed model.
\vspace{3pt}

\noindent\textbf{(iii) Deep Disguise Recognizer Network (DDRNET) \cite{ddrnetWvu}:} A team from West Virginia University, USA presented the DDRNET model. The name of the model was later changed to Deep Disguise Recognizer by the authors. Faces are cropped using the co-ordinates provided with the dataset, which is followed by pre-processing via whitening. An Inception network \cite{inception} along with Center loss \cite{center} is trained on the pre-processed images, followed by classification using a similarity metric. 
\vspace{3pt}

\noindent\textbf{(iv) DisguiseNet (DN) \cite{disguiseNetRopar}:} Submitted by a team from the Indian Institute of Technology, Ropar, DisguiseNet performs face detection using the facial co-ordinates provided with the dataset. A Siamese network is built using the pre-trained VGG-Face \cite{vgg}, which is fine-tuned with the DFW dataset. Cosine distance is applied for performing classification of the learned features.
\vspace{3pt}

\begin{figure}
\centering
\subfloat[Protocol-1]{\includegraphics[clip, trim=3cm 1cm 3cm 2.5cm, width=3.3in]{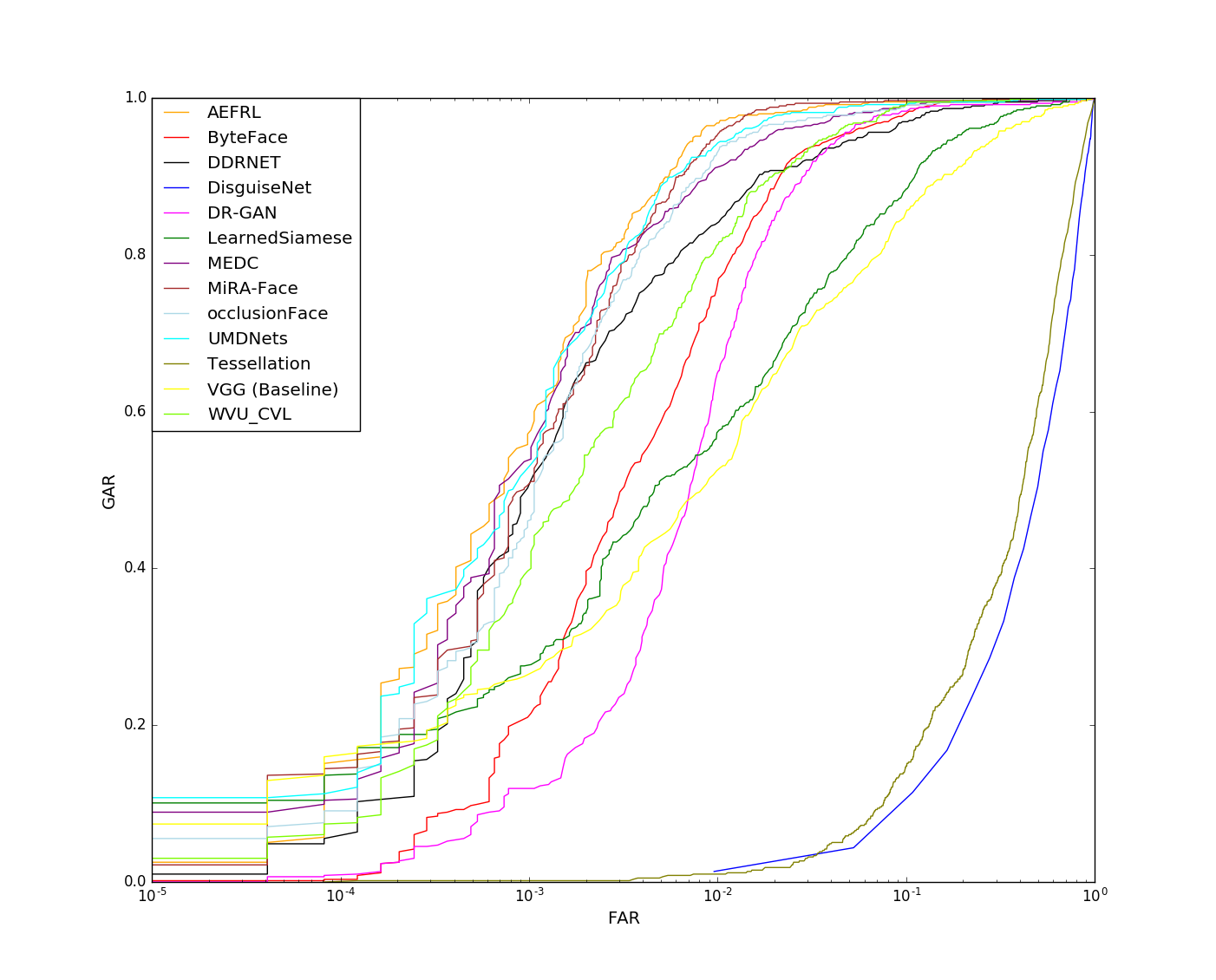}} \\
\subfloat[Protocol-2]{\includegraphics[clip, trim=2.5cm 1cm 3cm 2.5cm, width=3.3in]{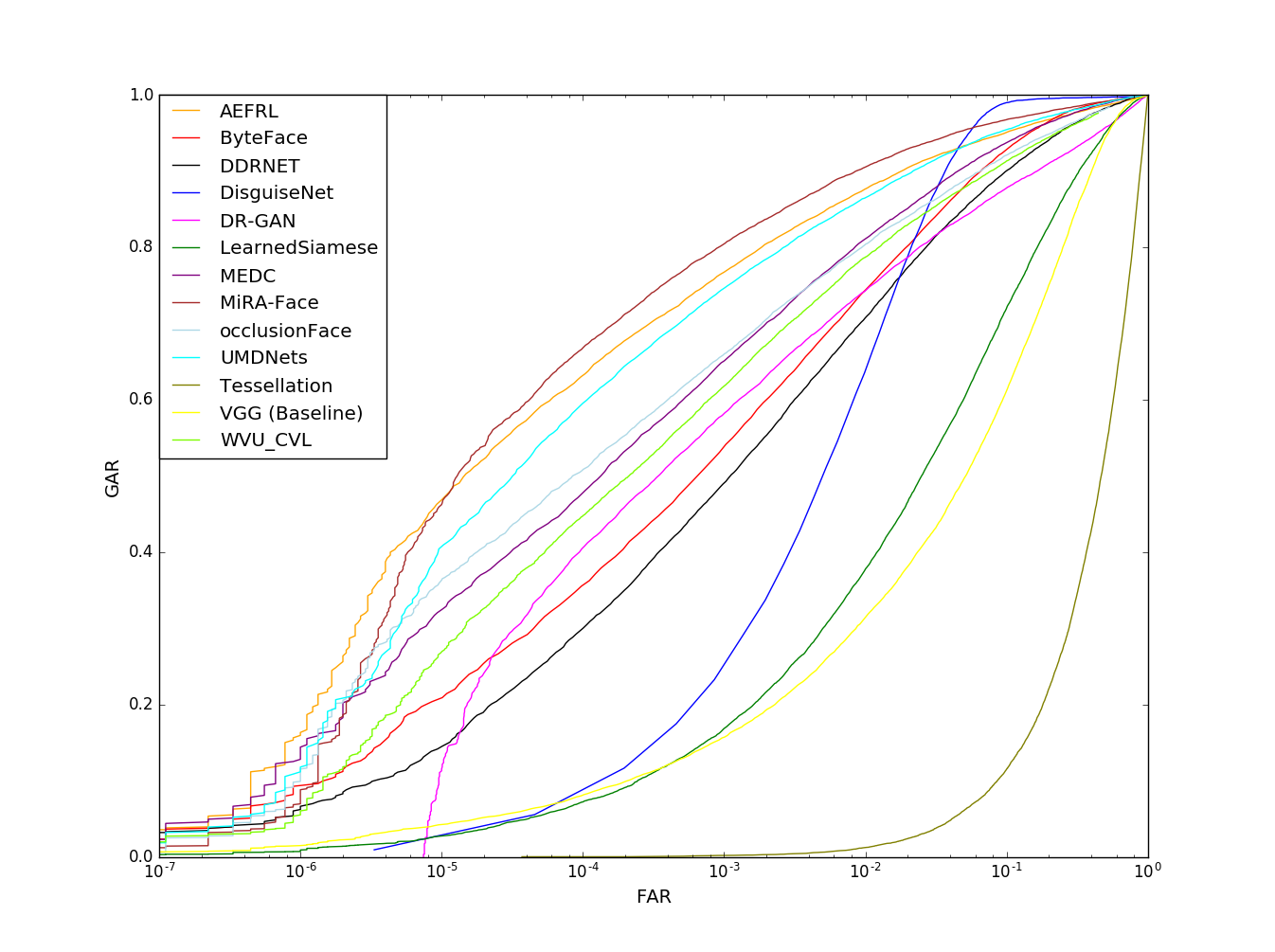}} \\
\subfloat[Protocol-3]{\includegraphics[clip, trim=2.5cm 1cm 3cm 2.5cm, width=3.3in]{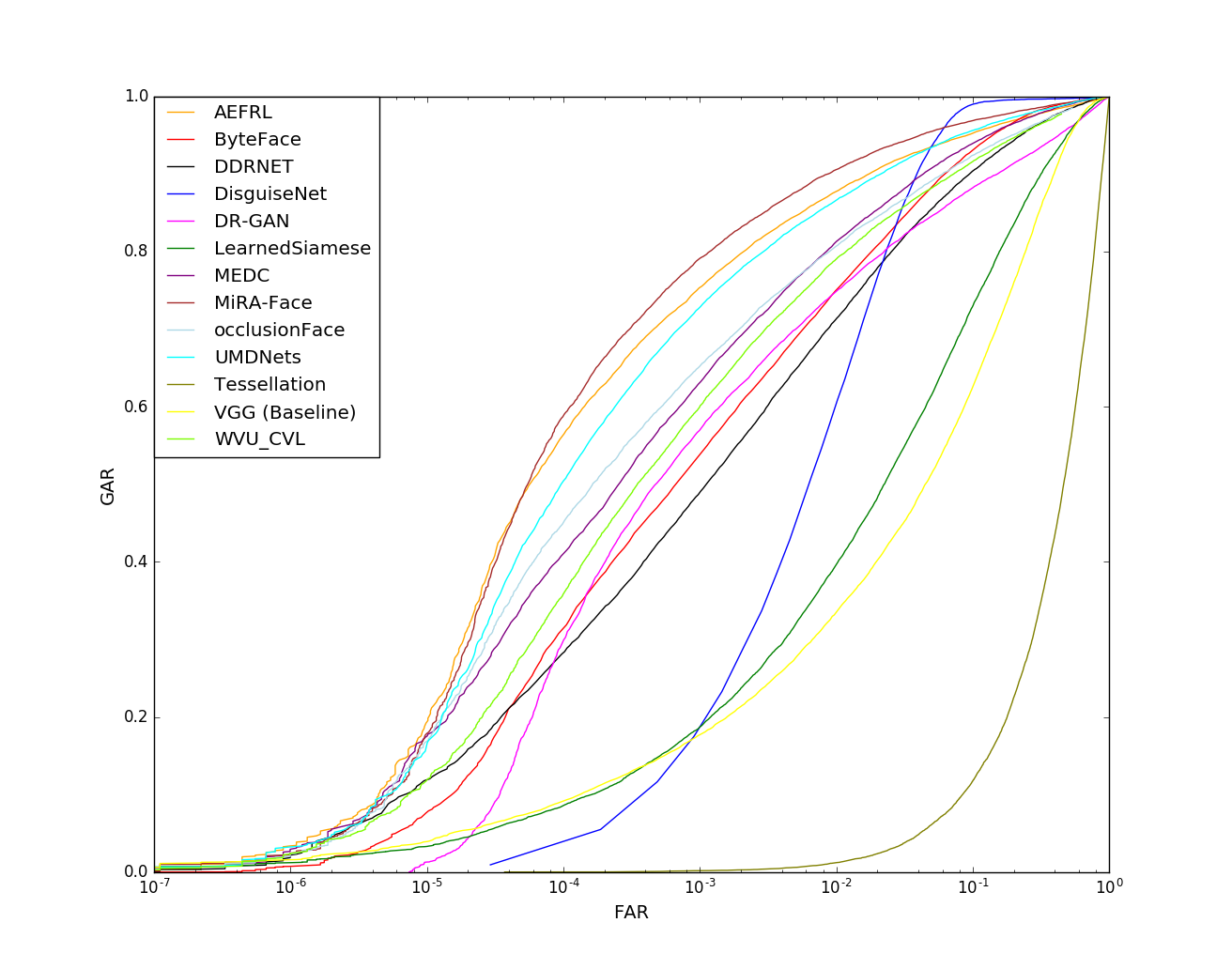}}
\caption{ROC curves of all participants along with the baseline results on protocol-1 (impersonation), protocol-2 (obfuscation), and protocol-3 (overall) of the DFW dataset.}
\label{fig:roc}
\end{figure}

\noindent\textbf{(v) DR-GAN:} Proposed by a team from Michigan State University, USA, the framework performs face detection and alignment on the input images using MT-CNN \cite{mtcnn}. This is followed by feature extraction using the Disentangled Representation learning-Generative Adversarial Network (DR-GAN) \cite{drGanArxiv}. Classification is performed using Cosine distance.
\vspace{3pt}

\noindent\textbf{(vi) LearnedSiamese (LS):} A team from the Computer Vision Center, Universitat Autònoma de Barcelona, Spain proposed LearnedSiamese. Facial co-ordinates provided with the dataset are used for performing face detection, followed by learning a Siamese Neural Network for disguised face recognition.
\vspace{3pt}

\noindent\textbf{(vii) Model Ensemble with Different CNNs (MEDC):} MEDC is proposed by a team from the Northeastern University, USA. Face detection is performed using MTCNN followed by 2-D alignment. An ensemble of three CNNs is used for performing the given task of disguised face recognition. The algorithm utilizes a Center face model \cite{center}, Sphere face model \cite{sphere}, and a ResNet-18 model \cite{resnet} trained on the MS-Celeb-1M dataset \cite{msCeleb}. Since MS-Celeb-1M dataset also contains images taken from the Internet, mutual exclusion is ensured with the test set of the DFW dataset. Classification is performed using Cosine distance for each network, the average of which is used for computing the final result.
\vspace{3pt}

\noindent \textbf{(viii) MiRA-Face \cite{miraFaceNTU}:} Submitted by a team from the National Taiwan University, MiRA-Face uses a combination of two CNNs for performing disguised face recognition. It treats aligned and unaligned images separately, thereby using a context-switching technique for a given input image. Images are aligned using the co-ordinates provided with the dataset along with MTCNN and Recurrent Scale Approximation (RSA) \cite{rsa}. Features learned by the CNNs are directly used for classification. Fig. \ref{fig:arch2} presents a diagrammatic representation of the proposed model.
\vspace{3pt}

\noindent \textbf{(ix) OcclusionFace:} A team from ZJU, China proposed the OcclusionFace framework. MT-CNN \cite{mtcnn} is used to perform face landmark detection and alignment based on five facial landmarks. ResNet-28 \cite{resnet} is used for performing classification. The model is first pre-trained on the CASIA Webface dataset \cite{webface} followed by fine-tuning on the DFW dataset. 
\vspace{3pt}

\noindent\textbf{(x) Tessellation:} Proposed by a team from Tessellate Imaging, India, Tessellation uses a Siamese network with triplet loss. Facial co-ordinates provided with the dataset are used for performing pre-processing, followed by training of the Siamese network. The final layer of the model learns a distance metric which returns a score between 0-1 for a given pair of images.
\vspace{3pt}

\noindent \textbf{(xi) UMDNets \cite{resnetUmd}:} Proposed by a team from University of Maryland, USA, its name was later modified to 'DCNN-based approach'. Face detection is performed by the All-in-One network \cite{allInOne}, followed by alignment using the detected keypoints. Feature extraction is performed using two networks, followed by independent score computation. Classification is performed by averaging the scores obtained via the two feature sets. Fig. \ref{fig:arch1}(b) presents the training and testing pipeline of the proposed model. 
\vspace{3pt}

\noindent \textbf{(xii) WVU\_CL:} Submitted by a team from West Virginia University, USA, WVU\_CL uses the face co-ordinates provided with the dataset along with MT-CNN \cite{mtcnn} for face alignment. The aligned images are provided to a CNN architecture for performing classification using a softmax classifier.

\subsection{Results}
Tables \ref{tab:protocol1}-\ref{tab:protocol3} and Fig. \ref{fig:roc} present the Receiver Operating Characteristic (ROC) curves of the above mentioned models for all three protocols. Along with the submissions, the performance of VGG-Face \cite{vgg} with Cosine distance is also tabulated as baseline. The performance of each model is reported in terms of Genuine Acceptance Rate (GAR) at 1\% False Acceptance Rate (FAR) and 0.1\% FAR. Results for each protocol are given in detail below:\\

\begin{table}
\centering
\caption{Verification accuracy (\%) of the participants and baseline performance on protocol-1 (impersonation).}
\begin{tabular}{|l|c|c|}
\hline
\multirow{2}{*}{\textbf{Algorithm}} & \multicolumn{2}{c|}{\textbf{GAR}} \\
\cline{2-3}
& \textbf{@1\%FAR} & \textbf{@0.1\%FAR} \\
\hline
\hline
\textbf{AEFRL} & \textbf{96.80} & \textbf{57.64}  \\
\hline
Baseline (VGG-Face) & 52.77 & 27.05\\
\hline
{ByteFace} & {75.53} & \textbf{55.11} \\
\hline
{DDRNET} & {84.20} & {51.26} \\
\hline
DenseNet + COST \tablefootnote {Not part of DFW competition\label{footnoteCOST}} & 92.1 & 62.2 \\
\hline
{DisguiseNet} & {1.34 \tablefootnote {GAR@0.95\%FAR}} & {1.34 \tablefootnote{The smallest FAR value is 0.95\%FAR for DisguiseNet.}}  \\
\hline
DR-GAN & 65.21 & 11.93\\
\hline
{LearnedSiamese} & {57.64} & {27.73}  \\
 \hline
{MEDC} & {91.26} & \textbf{55.46}  \\
\hline
\textbf{MiRA-Face} & \textbf{95.46} & {51.09} \\
\hline
OcclusionFace & 93.44 & 46.21 \\
\hline
{Tessellation} & {1.00} & {0.16}  \\
\hline
\textbf{UMDNets} & \textbf{94.28} & {53.27} \\
\hline
WVU\_CL & 81.34 & 40.00 \\
\hline
\end{tabular}
\label{tab:protocol1}
\end{table}

\noindent\textbf{Results on Protocol-1 (Impersonation):} Fig. \ref{fig:roc}(a) presents the ROC curves for all the submissions, and Table \ref{tab:protocol1} presents the GAR corresponding to two FAR values. It can be observed that for the task of impersonation, AEFRL outperforms other algorithms at both the FARs by achieving 96.80\% and 57.64\% at 1\% FAR and 0.1\%FAR, respectively. A difference of around 40\% is observed between the accuracies at both the FARs, which suggests that for scenarios having stricter authorized access, further improved performance is required. The second best performance is reported by MiRA-Face which presents a verification accuracy of 95.46\% and 51.09\%, respectively. 
At 0.1\%FAR, MEDC performs second best and achieves an accuracy of 55.46\%. All three algorithms utilize MT-CNNs for face detection and alignment before feature extraction and classification. 
\vspace{3pt}


\noindent \textbf{Results on Protocol-2 (Obfuscation):} Fig. \ref{fig:roc}(b) presents the ROC curves for the obfuscation protocol, and Table \ref{tab:protocol2} summarizes the verification accuracies for all the models, along with the baseline results. MiRA-Face achieves the best accuracy of 90.65\% and 80.56\% for the two FARs. It outperforms other algorithms by a margin of at least 2.8\% for GAR@1\%FAR and 2.5\% for GAR@0.1\%FAR. As compared to the previous protocol (impersonation), the difference in the verification accuracy at the two FARs is relatively less. While further improvement is required, however, this suggests that recognition systems suffer less in case of obfuscation, as compared to impersonation at stricter FARs.
\vspace{3pt}

\noindent \textbf{Results on Protocol-3 (Overall):} Table \ref{tab:protocol3} presents the GAR values of all the submissions, and Fig. \ref{fig:roc}(c) presents the ROC curves for the third protocol. As with the previous protocol, MiRA-Face outperforms other algorithms by a margin of at least around 3\%. An accuracy of 90.62\% and 79.26\% is reported by the model for 1\% and 0.1\%FAR. 

\begin{table}
\centering
\caption{Verification accuracy (\%) of the participants and baseline performance on protocol-2 (obfuscation).}
\begin{tabular}{|l|c|c|}
\hline
\multirow{2}{*}{\textbf{Algorithm}} & \multicolumn{2}{c|}{\textbf{GAR}} \\
\cline{2-3}
& \textbf{@1\%FAR} & \textbf{@0.1\%FAR} \\
\hline
\hline
\textbf{AEFRL} & \textbf{87.82} & \textbf{77.06}  \\
\hline
Baseline (VGG-Face) & 31.52 & 15.72\\
\hline
{ByteFace} & {76.97} & {21.51}  \\
\hline
DenseNet + COST \tablefootnotemark {footnoteCOST} & 87.1 & 72.1 \\
\hline
{DDRNET} & {71.04} & {49.28} \\
\hline
{DisguiseNet} & {66.32} & {28.99}  \\
\hline
DR-GAN & 74.56 & 58.31 \\
\hline
{LearnedSiamese}& {37.81} & {16.95} \\
\hline
{MEDC} & {81.25} & {65.14}  \\
\hline
\textbf{MiRA-Face} & \textbf{90.65} & \textbf{80.56} \\
\hline
OcclusionFace & 80.45 & 66.05 \\
\hline
{Tessellation} & {1.23} & {0.18}  \\
\hline
\textbf{UMDNets} & \textbf{86.62} & \textbf{74.69}  \\
\hline
WVU\_CL & 78.77 & 61.82 \\
\hline
\end{tabular}
\label{tab:protocol2}
\end{table}

\begin{table}
\centering
\caption{Verification accuracy (\%) of the participants and baseline performance on protocol-3 (overall).}
\begin{tabular}{|l|c|c|}
\hline
\multirow{2}{*}{\textbf{Algorithm}} & \multicolumn{2}{c|}{\textbf{GAR}} \\
\cline{2-3}
& \textbf{@1\%FAR} & \textbf{@0.1\%FAR} \\
\hline
\hline
\textbf{AEFRL} & \textbf{87.90} & \textbf{75.54} \\
\hline
Baseline (VGG-Face) & 33.76 & 17.73\\
\hline
{ByteFace} & {75.53} & {54.16} \\
\hline
DenseNet + COST \tablefootnotemark{footnoteCOST} & 87.6 & 71.5 \\
\hline
{DDRNET} & {71.43} & {49.08} \\
\hline
{DisguiseNet} &  {60.89} & {23.25}  \\
\hline
DR-GAN & 74.89 & 57.30 \\
\hline
{LearnedSiamese}& {39.73} & {18.79} \\
\hline
{MEDC} & {81.31} & {63.22} \\
\hline
\textbf{MiRA-Face} & \textbf{90.62} & \textbf{79.26} \\
\hline
OcclusionFace & 80.80 & 65.34 \\
\hline
{Tessellation} & {1.23} & {0.17}  \\
\hline
\textbf{UMDNets} & \textbf{86.75} & \textbf{72.90} \\
\hline
WVU\_CL & 79.04 & 60.13 \\
\hline
\end{tabular}
\label{tab:protocol3}
\vspace{-10pt}
\end{table}

\begin{figure}[h]
\centering
\includegraphics[width=3.4in]{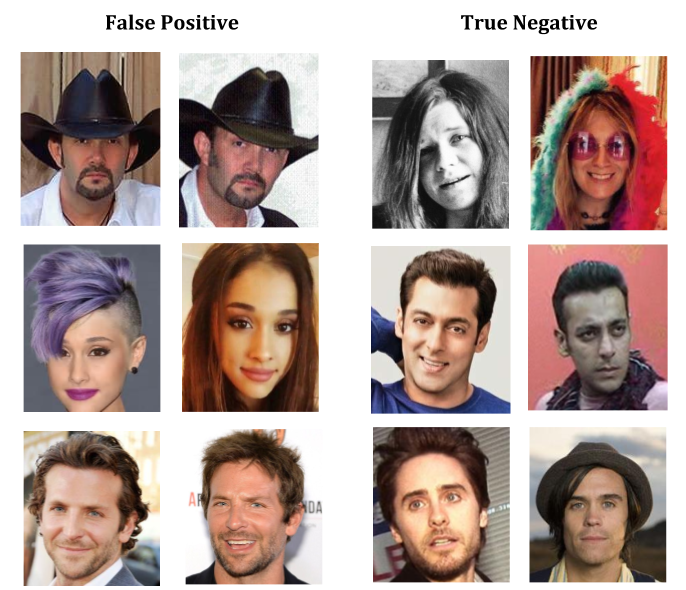}
\caption{Sample False Positive and True Negative pairs reported by a majority of submissions for protocol-1 (impersonation). False Positive refers to the case where an algorithm incorrectly classifies a pair as genuine, and True Negative refers to the case where two samples of different identities are correctly classified as imposters.}
\label{fig:Image_pair_case1}
\vspace{-10pt}
\end{figure}

\begin{figure}[h]
\centering
\includegraphics[width=3.4in]{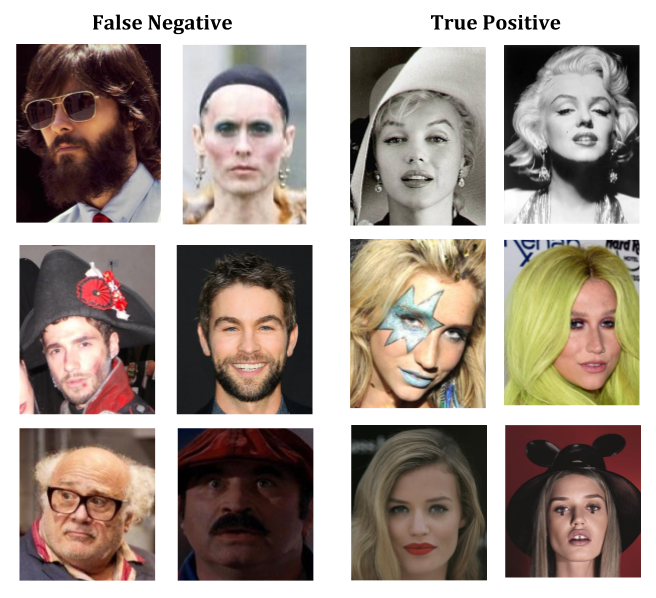}
\caption{Sample False Negative and True Positive pairs reported by a majority of submissions for protocol-2 (obfuscation). False Negative refers to the case where a pair of images are incorrectly classified as an imposter pair, while True Positive refers to the scenario where a pair of images are correctly classified as a genuine pair.}
\label{fig:Image_pair_case2}
\vspace{-10pt}
\end{figure}

Other than the DFW competition submissions, Suri \textit{et al.} \cite{suriBtas18} proposed a novel  COST (Color (CO), Shape (S), and Texture (T)) based framework for performing disguised face recognition. COST learns different dictionaries for Color, Shape, and Texture, which are used for feature extraction, along with the deep learning based model, DenseNet \cite{densenet}. Final output is computed via classifier level fusion of the deep learning and dictionary learning models. The performance of the proposed DenseNet + COST algorithm has also been tabulated in Tables \ref{tab:protocol1} - \ref{tab:protocol3}.

\begin{figure*}
\centering
\subfloat[\textit{Easy} Genuine Samples]{\includegraphics[width=6in]{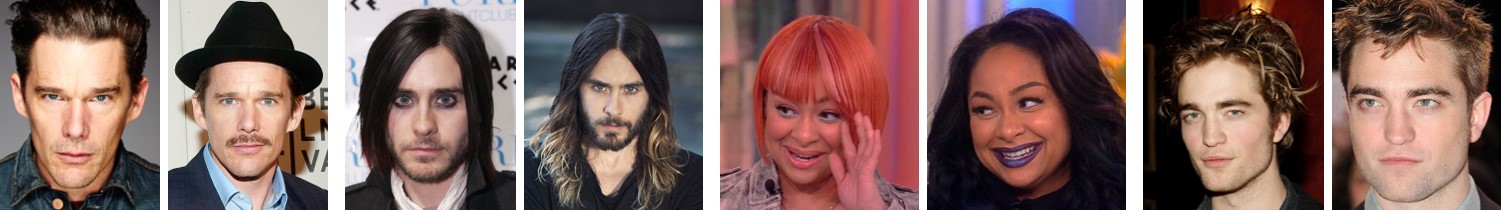}} \\
\subfloat[\textit{Hard} Genuine Samples]{\includegraphics[width=6in]{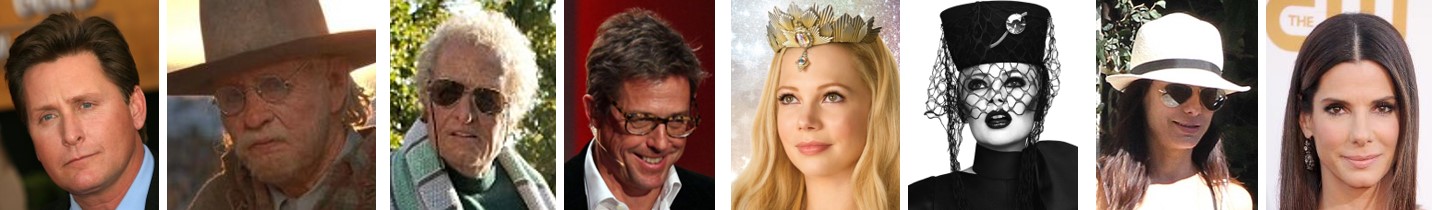}} \\
\subfloat[\textit{Hard} Imposter Samples]{\includegraphics[width=6in]{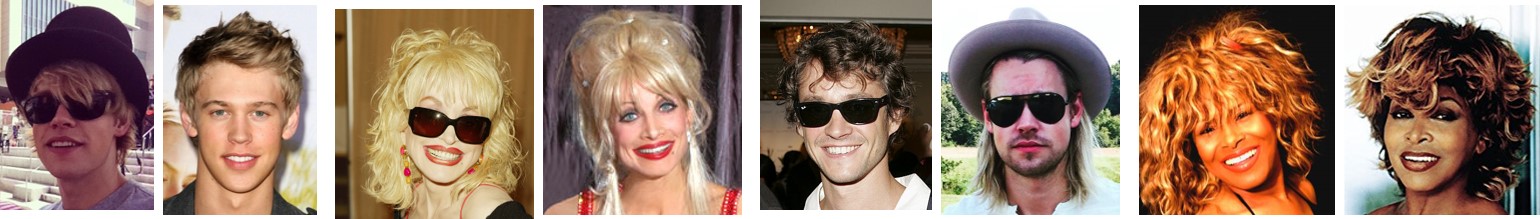}}
\caption{Sample \textit{easy} and \textit{hard} pairs of the DFW dataset.}
\label{fig:gbu}
\vspace{-5pt}
\end{figure*}

\begin{table*}
\centering
\caption{Number of \textit{easy}, \textit{medium}, and \textit{hard} pairs for 1\% and 0.1\% FAR. TP and TN refer to True Positive and True Negative, respectively.}
\vspace{-5pt}
\label{tab:gbu}
\begin{tabular}{|c|c|c|c||c|c|c||c|c|c|c|}
\hline
\multirow{3}{*}{\textbf{FAR}} & \multicolumn{9}{c|}{\textbf{Number of}} \\
\cline{2-10}
 & \multicolumn{3}{c||}{\textbf{Easy}} & \multicolumn{3}{c||}{\textbf{Medium}} & \multicolumn{3}{c|}{\textbf{Hard}} \\
\hline
& \textbf{Genuine (TP)} & \textbf{Imposter (TN)} & \textbf{Total} & \textbf{Genuine (TP)} & \textbf{Imposter (TN)} & \textbf{Total} & \textbf{Genuine (TP)} & \textbf{Imposter (TN)} & \textbf{Total} \\
\hline
\hline
1\% & 11,544 & 8,878,599 & 8,890,143 & 789 & 106,398 & 107,187 & 1,564 & 67,435 & 68,999\\
\hline
0.1\% & 9,461 & 9,034,109 & 9,043,570 & 1,138 & 11,534 & 12,672 & 3,298 & 6,789 & 10,087\\
\hline
\end{tabular}
\vspace{-10pt}
\end{table*}

Figs. \ref{fig:Image_pair_case1} - \ref{fig:Image_pair_case2} demonstrate sample images of the DFW dataset correctly classified or misclassified by almost all the submissions. Fig. \ref{fig:Image_pair_case1} presents False Positive and True Negative samples for protocol-1 (impersonation). Upon analyzing the False Positive samples, it can be observed that all pairs have similar lower face structure, which might result in algorithms incorrectly classifying them as the same subject. Moreover, external disguises such as the cowboy hat (first pair) might also contribute to the misclassification. For protocol-2 (obfuscation), Fig. \ref{fig:Image_pair_case2} presents sample False Negative and True Positive pairs common across almost all submissions. It is interesting to observe that in the False Negative pairs, disguise results in modification of face structure and textural properties. Coupled with obfuscation of face and pose variations, the problem of disguised face recognition is rendered further challenging.

\section{Degree of Difficulty: Easy, Medium, and Hard}
\label{sec:deg}
In order to further analyze the DFW dataset, and study the problem of disguised faces in the wild, the DFW dataset has been partitioned into three sets: (i) easy, (ii) medium, and (iii) hard. The \textit{easy} partition contains pairs of face images which are relatively easy to classify by a face recognition system, the \textit{medium} set contains pairs of images which can be matched correctly by a majority of face recognition systems, while the \textit{hard} partition contains image pairs with high matching difficulty. In literature, a similar partitioning was performed for the Good, the Bad, and the Ugly (GBU) face recognition challenge \cite{gbu}, where a subset of FRVT 2006 competition data \cite{frvt} was divided into the three sets. The GBU challenge contained data captured over an academic year, in constrained settings with frontal face images having minimal pose or appearance variations. This section analyzes the proposed DFW dataset containing data captured in unconstrained scenarios with variations across disguise, pose, illumination, age, and acquisition device.

For the DFW dataset, the top-3 performing algorithms of the DFW competition have been used for partitioning the dataset, that is, AERFL, MiRA-Face, and UMDNets. The performance of the three algorithms is used for dividing the test set of the DFW dataset into three components: (i) easy, (ii) medium, and (iii) hard. \textit{Easy} samples correspond to those pairs which were correctly classified by all three algorithms, and are thus easy to classify. \textit{Medium} samples were correctly classified by any two algorithms, while the \textit{hard} samples were correctly classified by only one algorithm, or mis-classified by all the algorithms, and thus are the most challenging component of the dataset. It is ensured that the partitions are disjoint, and samples belonging to one category do not appear in another category.

\begin{figure}
\centering
\includegraphics[width=3.6in]{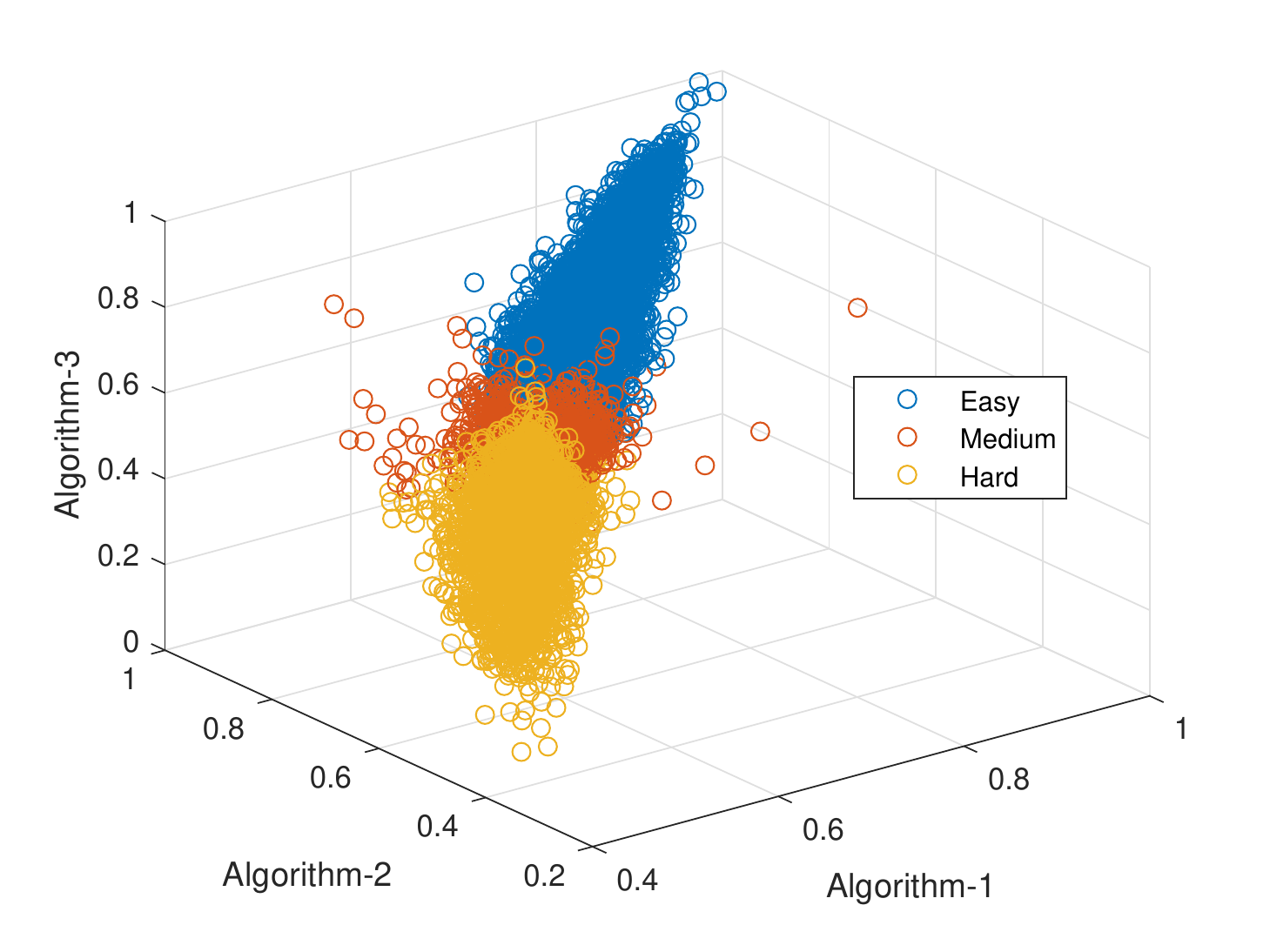}
\caption{Score distribution of the genuine pairs at 0.01\% FAR, in terms of three levels of difficulty: easy, medium, and hard.}
\label{fig:scores}
\vspace{-10pt}
\end{figure}

Table \ref{tab:gbu} presents the number of easy, medium, and hard pairs at different False Accept Rates of 1\% and 0.1\%. At 1\%FAR, 11,544 genuine pairs are correctly classified as True Positive, while 8,878,599 imposter pairs are correctly classified as True Negative by all three techniques. This results in a total of 8,890,143 \textit{easy} pairs, signifying that the total number of \textit{easy} samples are highly dominated by the imposter pairs. In comparison, at 0.1\%FAR, the total number of \textit{easy} pairs increase to 9,043,570. It is interesting to observe that this increase is primarily due to the increased number of \textit{easy} imposters at the lower FAR. Since at lower FARs, more pairs are classified as imposters, it leads to an increased number of \textit{easy} pairs. Intuitively, at a stricter threshold of 0.1\%FAR, one would expect the number of \textit{easy} genuine samples to reduce. This trend is observed in Table \ref{tab:gbu}, where the number of genuine pairs reduces from 11,544 at 1\%FAR to 9,461 at 0.1\%FAR. 

The opposite trend is observed for the \textit{hard} partition, where the total number of \textit{hard} pairs reduces at 0.1\%FAR, as compared to 1\%FAR, however, the number of genuine samples increases. The last three columns of Table \ref{tab:gbu} can be analyzed in order to observe this effect. At 1\%FAR, the number of \textit{hard} genuine pairs, that is, samples which are classified correctly by at most one algorithm is 1,564, while at 0.1\%FAR it is 3,298. This implies that at a stricter FAR of 0.1\%, more genuine samples were misclassified by all three algorithms. Parallely, the number of \textit{hard} imposter samples drops from 67,435 to 6,789 at a lower FAR. A similar trend is observed for the \textit{medium} partition, wherein a total of 107,187 and 12,672 samples were correctly classified by any two algorithms at 1\% and 0.1\%FAR, respectively. 

Fig. \ref{fig:scores} presents the score distribution of the genuine samples across the three categories of \textit{easy}, \textit{medium}, and \textit{hard} at 0.1\%FAR. The \textit{easy} and \textit{hard} samples occupy opposite ends of the distribution, while the \textit{medium} category corresponds to a dense block between the two. Fig. \ref{fig:gbu} presents sample \textit{easy} and \textit{hard} pairs of the DFW dataset at 0.1\% FAR. The first row corresponds to \textit{easy} genuine pairs, that is, genuine pairs correctly classified by all three top performing algorithms. Most of these pairs contain images with no pose variations ((i)-(ii)) or \textit{similar} pose variations across images of the pairs ((iii) -(iv)). It can also be observed that most of these pairs are of normal and validation images of the dataset, with minimal or no disguise variations. Images which involve disguise in terms of hair variations or hair accessories with minimal change in the face region are also viewed as \textit{easy} pairs by the algorithms. Since in such cases, the face region remains unchanged, algorithms are often able to correctly classify such samples with ease. This observation is further substantiated by the \textit{hard} genuine samples (Fig. \ref{fig:gbu}(b)). Most of the samples which were not correctly classified by any of the top algorithms contain occlusions in the face region. A large majority of genuine samples misclassified have occlusions near the eye region. All the pairs demonstrated in Fig. \ref{fig:gbu}(b) have at least one sample with occluded eye region. Effect of occlusion can also be observed in the \textit{hard} imposter samples (Fig. \ref{fig:gbu}(c)), that is, imposters which were not correctly classified by either of the top-3 performing algorithms. Large variations due to heavy make-up, similar hair style or accessories, coupled with covariates of pose, occlusion, illumination, and acquisition device further make the problem challenging. It is our belief that in order to develop robust face recognition systems invariant to disguises, research must focus on addressing the \textit{hard} pairs, while ensuring high performance on the \textit{easy} pairs as well.

\section{Conclusion} 
This research presents a novel Disguised Faces in the Wild (DFW) dataset containing 11,157 images pertaining to 1,000 identities with variations across different disguise accessories. A given subject may contain four types of images: normal, validation, disguised, and impersonator. Out of these, normal and validation images are non-disguised frontal face images. Disguised images of a subject contain genuine images of the same subject with different disguises. Impersonator images correspond to images of different people who try to impersonate (intentionally or unintentionally) another subject. To the best of our knowledge, this is the first disguised face dataset to provide impersonator images for different subjects. Three evaluation protocols have been presented for the proposed dataset, along with the baseline results. The proposed dataset has also been analyzed in terms of three degree of difficulty: (i) easy, (ii) medium, and (iii) hard. The dataset was released as part of the DFW competition held in conjunction with the First International Workshop on DFW at CVPR'18. Details regarding the submissions and their performance evaluation has also been provided. The dataset is made publicly available for research, and it is our hypothesis that it would help facilitate research in this important yet less explored domain of face recognition. 

\section{Acknowledgement}
M. Singh, R. Singh, and M. Vatsa are partially supported through Infosys CAI at IIIT-Delhi. The authors acknowledge V. Kushwaha for his help in data collection. Rama Chellappa was supported by the Office
of the Director of National Intelligence (ODNI), Intelligence Advanced Research Projects Activity (IARPA), via IARPA R\&D Contract No. 2014-14071600012. The views and conclusions contained herein are those of the authors and should not be interpreted as necessarily representing the official policies or endorsements, either expressed or implied, of the ODNI, IARPA, or the U.S. Government. The U.S. Government is authorized to reproduce and distribute reprints for Governmental purposes notwithstanding any copyright annotation thereon.



{\small
\bibliographystyle{IEEEtran}
\bibliography{egbib}
}

\end{document}